\documentclass[aps,pre,twocolumn,groupedaddress]{revtex4}
\usepackage{graphicx}
\usepackage{amsmath}
\usepackage[utf8]{inputenc}

\usepackage{algcompatible}
\usepackage{xcolor}

\begin{document}
\title{Optimizing Memory in Reservoir Computers }
\author{T. L. Carroll}
\email{Thomas.Carroll@nrl.navy.mil}
\affiliation{US Naval Research Lab, Washington, DC 20375}

\date{\today}

\begin{abstract}
A reservoir computer is a way of using a high dimensional dynamical system for computation. One way to construct a reservoir computer is by connecting a set of nonlinear nodes into a network. {Because the network creates feedback between nodes, the reservoir computer has memory. If the reservoir computer is to respond to an input signal in a consistent way (a necessary condition for computation), the memory must be fading;} that is, the influence of the initial conditions fades over time. How long this memory lasts is important for determining how well the reservoir computer can solve a particular problem. In this paper I describe ways to vary the length of the fading memory in reservoir computers. Tuning the memory can be important to achieve optimal results in some problems; too much or too little memory degrades the accuracy of the computation. 
\end{abstract}

\maketitle

{\bf
The theory of computing by means of dynamical systems states that computing requires memory, information transmission and a way for stored and transmitted information to interact. Reservoir computers are one way to arrange a dynamical system for computation. To implement a reservoir computer, a high dimensional dynamical system is driven by one or more signals from a system to be analyzed. Typically the dynamical system is constructed by connecting a large number of nonlinear nodes in a network. The network contains feedback paths, so that the reservoir computer is a dynamical system. Training of a reservoir computer proceeds by fitting output time series signals from the reservoir computer to a training signal which has some relation to the input signal. Unlike a standard recurrent neural network, the connections between nodes in a reservoir computer are not changed, making training much faster for a reservoir computer. There are several dynamical characteristics of a reservoir computer that must be tuned to fit the particular problem being solved; among those is the length of the fading memory. In this paper I show some strategies for adjusting the memory length.
}

\section{Introduction}
It has been stated that computation by means of a dynamical system requires that the system have memory and be able to transmit signals \cite{langton1990}. Memory is necessary for computation, but the amount of memory depends on the particular computation. One way to arrange a dynamical system to do computation is as a reservoir computer, also known as an echo state machine or a liquid state machine \cite{jaeger2001, natschlaeger2002}. In a reservoir computer, a set of nonlinear nodes is connected into a network. The nodes are driven by a common input signal, and the time series responses of all of the nodes are recorded. In some ways the reservoir computer resembles a recurrent neural network, but unlike a neural network, the connections between nodes in a reservoir computer are not changed. Rather, training proceeds by fitting the node outputs to a training signal. The fit coefficients are the output of the training procedure. The fact that the connections between nodes are never changed means that reservoir computers may be constructed from analog hardware where altering the connections between nodes may be difficult. Keeping a fixed network also removes constraints on the node activation function imposed by training requirements, so many different nonlinear devices may be used as reservoir computer nodes.

The memory in a reservoir computer must be fading memory. Boyd and Chua \cite{boyd1985} define fading memory in a way that resembles the definition of the largest Lyapunov exponent for a dynamical system.  

Examples of reservoir computers so far include photonic systems \cite{appeltant2011,larger2012, van_der_sande2017}, analog circuits \cite{schurmann2004}, mechanical systems \cite{dion2018} and  field programmable gate arrays \cite{canaday2018}. Many other examples are included in the review paper \cite{tanaka2019}, which describes hardware implementations of reservoir computers that  are very fast, and yet consume little power, while being small and light. Reservoir computers have been shown to be useful for solving a number of problems, including reconstruction and prediction of chaotic attractors \cite{lu2018,zimmerman2018,antonik2018,lu2017,jaeger2004}, recognizing speech, handwriting or other images \cite{jalavand2018} or controlling robotic systems \cite{lukosevicius2012} . Reservoir computers have also been used to better understand the function of neurons in the brain \cite{stoop2013}. Several groups have been using theory to better understand reservoir computers; in  \cite{hart2020}, the authors show that there is a positive probability that a reservoir computer can be an embedding of the driving system, and therefore can predict the future of the driving system within an arbitrary tolerance, while Grigoryeva et al. \cite{grigoryeva2021} show conditions under which a reservoir computer can be in strong generalized synchronization with the driving system.  Lymburn et al. \cite{lymburn2019} study the relation between generalized synchronization and reconstruction accuracy, while Herteux and R{\"a}th examine how the symmetry of the activation function affects reservoir computer performance \cite{herteux}. 

Because the network in a reservoir computer is not trained, there are a number of parameters that must be chosen ahead of time to optimize the performance of the reservoir computer. Motivated by work such as that of Langton \cite{langton1990}, some researchers have studied ways to adjust the memory of a reservoir computer to the problem being solved.  It has been shown that in some situations a reservoir computer can have too much memory \cite{carroll2020b}. In a different example that demonstrated that only a short memory may be necessary, Gauthier et al. \cite{gauthier2021} were able to predict the signals from a Lorenz chaotic system with a series approximation that had a memory of only one time step. Luko{\v s}evi{\v c}ius \cite{lukosevicius2012a} points out that adjusting the memory capacity of the reservoir computer for the particular problem is necessary. In \cite{verstraeten2010} the authors show how bias, input scaling and spectral radius affect the memory capacity of a reservoir computer using tanh nodes. 

 {\subsection{Is Memory Capacity a Useful Statistic?}
In addressing memory, it is necessary to define what is meant by "memory". The Boyd and Chua work \cite{boyd1985} gives a definition, but this definition is not the one most commonly used in discussing memory capacity. The most widely used definition, which measures the ability to predict previous values of a white noise signal, was introduced in \cite{jaeger2002}. This definition does not depend on the actual signal driving the reservoir computer, but in a nonlinear system the dynamics depends both on the system and the driving signal, so two other memory measures are introduced here; one which like the Jaeger memory capacity depends on the change in correlation over time within a reservoir computer but using the actual driving signal, and another definition which resembles the Boyd and Chua definition.}

 {While the real measure of performance in a reservoir computer is how accurately it can compute or classify a desired signal, some papers on reservoir computing only examine memory capacity without also measuring testing error. Still, in some cases memory by itself can be a useful statistic, so concentrating on optimizing memory can be helpful in improving the performance of a reservoir computer. The standard model of memory in computation  \cite{langton1990} states that longer memory leads to higher computational capacity, and in many cases this is true. On the other hand, \cite{carroll2020b} showed that a reservoir computer can have too much memory. These dynamical effects can be measured by comparing the Lyapunov exponent spectrum of the reservoir computer to the Lyapunov spectrum of the driving system, but in an analog reservoir computer the equations describing the reservoir may not be known. In addition, there is a range of Lyapunov exponents for the reservoir computer, and they may all be important. Memory capacity can be a way to describe the reservoir computer dynamics using only one number, and two of the three memory statistics described in this work do not require knowledge of the dynamical system.}

\section{Reservoir Computer}
\label{rescomp}
The reservoir computer I use in this work is based on a tanh function:
\begin{equation}
\label{tanhnode}
{\bf{R}}\left( {n + 1} \right) = g\tanh \left( {{\bf{AR}} + \varepsilon s\left( n \right)} \right)
\end{equation}
 where ${\bf R}$ is the vector of reservoir variables, ${\bf A}$ is the adjacency matrix and $s(n)$ is the input signal. All the entries in the adjacency matrix were set to values drawn from a Gaussian random distribution with mean 0 and standard deviation 1, and then ${\bf A}$ was renormalized to have a spectral radius of 1.  For the work in this paper, ${\bf A}$ had $M=100$ nodes. Some papers using a tanh node add a bias within the tanh function, but in order to reduce the number of parameters, I set the bias to zero. {For small signals the tanh function is approximately linear, so a bias will move small signals into the nonlinear region of the tanh. The effects of the nonlinearity may also be increased by increasing the input constant $\varepsilon$. }
 
  In the training stage, the reservoir computer is driven with the input signal $s(n)$ to produce the reservoir computer output signals $r_i(n)$. The first 1000 points from the $r_i(n)$ time series are discarded and the next 10,000 points are used to fit the training signal $f(n)$. The tanh nonlinearity is asymmetric about zero, so in order to add terms symmetric about zero, a matrix is constructed from the reservoir signals as
\begin{equation}
\label{fitmat}
\Omega  = \left[ {\begin{array}{*{20}{c}}
{{r_1}\left( 1 \right)}& \cdots &{{r_M}\left( 1 \right)}&{r_1^2\left( 1 \right)}& \cdots &{r_M^2\left( 1 \right)}\\
{{r_1}\left( 2 \right)}&{}&{{r_M}\left( 2 \right)}&{r_1^2\left( 2 \right)}&{}&{r_M^2\left( 2 \right)}\\
 \vdots &{}& \vdots &{}&{}& \vdots \\
{{r_1}\left( N \right)}& \cdots &{{r_M}\left( N \right)}&{r_1^2N}& \cdots &{r_M^2\left( N \right)}
\end{array}} \right]
\end{equation}
where the reservoir has $M$ nodes and the time series have $N$ points. The fit to the training signal is 
\begin{equation}
\label{train_fit}
{h(t)} ={\Omega } {{\bf C}}
\end{equation}
where ${\bf C}$ is a vector of training coefficients. The fit is usually done by a ridge regression to avoid overfitting

 The training error is 
\begin{equation}
\label{train_err}
{\Delta _{RC}} = {{\left\langle {f\left( n \right) - h\left( n \right)} \right\rangle } \mathord{\left/
 {\vphantom {{\left\langle {f\left( n \right) - h\left( n \right)} \right\rangle } {\left\langle {f\left( n \right)} \right\rangle }}} \right.
 \kern-\nulldelimiterspace} {\left\langle {f\left( n \right)} \right\rangle }}
 \end{equation}
 where $\left\langle \; \right\rangle $ indicates a standard deviation. 
 
 In the testing stage, a new input signal $\tilde s\left( n \right)$ is generated from the same dynamical system that generated $s(n)$, but with different initial conditions. The corresponding test signal is $\tilde f\left( n \right)$. The input signal $\tilde s\left( n \right)$ drives the same reservoir to produce the output signals ${\tilde r_i}\left( n \right)$, which are arranged in a matrix $\tilde \Omega$. The testing error is
 \begin{equation}
 \label{test_err}
 {\Delta _{tx}} = {{\left\langle {\tilde f\left( n \right) - \tilde \Omega {\bf{C}}} \right\rangle } \mathord{\left/
 {\vphantom {{\left\langle {\tilde f\left( n \right) - \tilde \Omega {\bf{C}}} \right\rangle } {\left\langle {\tilde f\left( n \right)} \right\rangle }}} \right.
 \kern-\nulldelimiterspace} {\left\langle {\tilde f\left( n \right)} \right\rangle }}
\end{equation}
where the coefficient vector ${\bf C}$ was found in the training stage.
 
 \section{Lyapunov Exponents}
 The Lyapunov exponents were estimated by the Gram-Schmidt method, as described in \cite{Parker:1989}. To estimate the largest $n_{\lambda}$ Lyapunov exponents for the reservoir computer, an $M \times M$ variational matrix $\Theta$ was created. The matrix $\Theta$ was initially set to the identity. The variational matrix was propagated in time as
\begin{equation}
\label{vareq}
{ \Theta \left(k+1 \right) = {D_r}f\left( {{\bf{r}}\left( k \right)} \right)\Theta \left( k \right)}
\end{equation}
where ${D_r}f\left( {{\bf{r}}\left( k \right)} \right)$ is the Jacobian of eq. (\ref{tanhnode}).

A perturbation $\delta{\bf r}(0)$ of the initial condition ${\bf r}(0)$ evolves as $\delta {\bf{r}}\left( 0 \right) = \Theta \left( k \right)\delta {\bf{r}}\left( 0 \right)$. To find the largest $n_{\lambda}$ Lyapunov exponents for the reservoir computer, $\delta{\bf r}(0)$ is initially set to a random $M \times n_{\lambda}$ matrix and the columns are orthonormalized. To find the Lyapunov exponents for the different directions in the reservoir computer phase space, after each time step $k$ the columns of $\delta{\bf r}(k)$ are orthogonalized using the Gram-Schmidt procedure. The Lyapunov exponents at this time step are obtained from the logarithms of the norms of the orthogonal vectors. The variational equation is then propagated to the next time step, $k+1$. If the norm of the variational matrix becomes too large or too small, so that numerical accuracy is affected, the variational matrix is reset to a random orthonormal matrix.
 
 The Gram-Schmidt method is not as accurate as the QR decomposition method \cite{eckmann1985}, but the full Lyapunov exponent spectrum of the reservoir computer was not required. In this case, only the largest Lyapunov exponent was used, so the Gram-Schmidt method could be used to avoid decomposing very large matrices.
 
\section{Measures of Memory}
\label{memstat}
\subsection{Memory Capacity}
\begin{figure}
\centering
\includegraphics[scale=0.8]{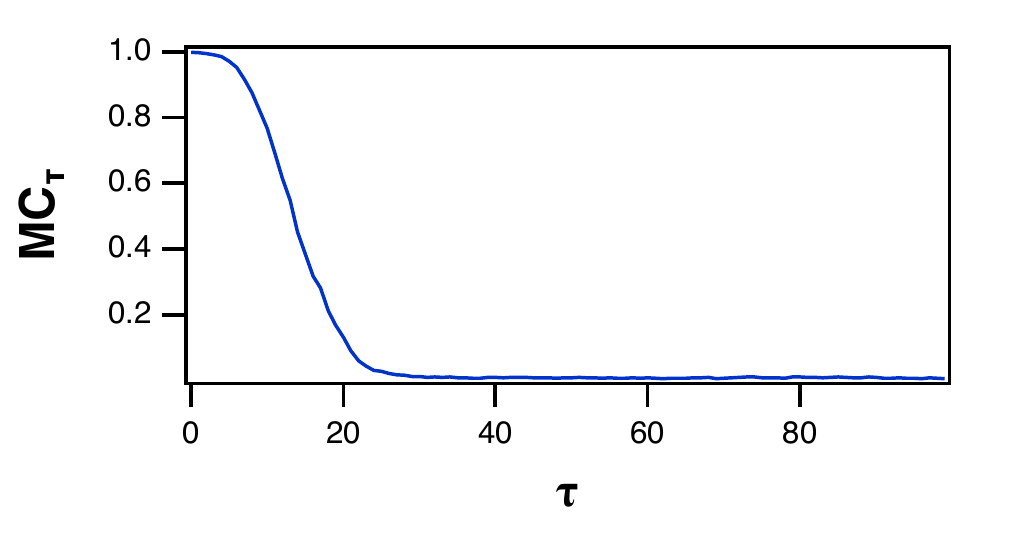} 
  \caption{ \label{memtau} Memory capacity $MC_{\tau}$ as defined in eq. (\ref{memdel}) for the tanh reservoir as a function of the delay $\tau$. For this plot, $\varepsilon=1$ and $g=1$.}
  \end{figure}

The standard calculation of memory in reservoir computers is the memory capacity introduced in \cite{jaeger2002}. {This memory capacity is a measure of how well the reservoir computer can predict previous values of the input signal, with the quality of fit being measured by the cross correlation between the input signal $s(n-\tau)$ and the reservoir computer fit to this signal.  To avoid confusing correlations induced by the reservoir computer with correlations in the input signal}, the reservoir input signal $s(n)$ is a random noise signal- in this work, the noise is Gaussian with a standard deviation of 1. The memory capacity as a function of delay is calculated as
\begin{equation}
\label{memdel}
{\rm{M}}{{\rm{C}}_\tau } = \frac{{\sum\limits_{n = 1}^N {{{\left( {\left[ {s\left( {n - \tau } \right) - \bar s} \right]\left[ {{h_\tau }\left( n \right) - \overline {{h_\tau }} } \right]} \right)}^2}} }}{{\sum\limits_{n = 1}^N {{{\left[ {s\left( {n - \tau } \right) - \bar s} \right]}^2}\sum\limits_{n = 1}^N {{{\left[ {{h_\tau }\left( n \right) - \overline {{h_\tau }} } \right]}^2}} } }}
\end{equation}

with $N=10 000$ and the overbar indicator indicates the mean.  The signal $h_{\tau}(n)$ is the fit of the reservoir signals $r_i(n)$ to the delayed input signal $s(n-\tau)$. The extra squared terms used in the fitting matrix $\Omega$ in eq. (\ref{fitmat}) do not add to the memory capacity, so they are not included in the fit for the purpose of calculating memory capacity. The memory capacity as a function of delay, $MC_{\tau}$, is plotted in figure \ref{memtau}.

The total memory capacity is
\begin{equation}
\label{memcap}
{\rm{MC}} = \sum\limits_{\tau  = 1}^{{\tau _{\max }}} {{\rm{M}}{{\rm{C}}_\tau }} 
\end{equation}
where $\tau_{max}$ was set to 100 because at that value ${\rm MC}_{\tau}$ was small. 

There are some drawbacks to this definition of memory. Changing the input signal for a nonlinear system will change its properties- even changing the amplitude of the random input signal will change the effect of the nonlinearities in the reservoir computer, so the memory capacity as defined in eq. (\ref{memcap}) may not be a true reflection of the memory of the reservoir computer. Also, the memory capacity calculation requires that one fit a signal, but memory capacity is often used as an independent metric to define how well the reservoir computer will fit signals. Characteristics of the reservoir computer that are independent of the memory but lead it to be better or worse at reproducing signals can affect the memory capacity estimate. Because of these drawbacks, two alternate estimates of memory are used in this paper.

\subsection{Norm of the Variation}
Inubushi and Yoshimura \cite{inubushi2017} use the variational equation for the reservoir computer to prove that the memory for a nonlinear system is less than the memory for a linear system. This statistic also is similar to the Boyd and Chua definition of memory \cite{boyd1985}. Following their example, the variational equation may be used to measure the amount of memory in a reservoir computer. If the initial perturbation is $\delta_0$, the perturbation at $n$ steps later for a  nonlinear system is, 
 \begin{equation}
 \label{nljac}
 \delta_n = \left[ {\prod\limits_{j = 0}^n {{D_r}f\left( {{\bf{r}}\left( j \right)} \right)} } \right]{\delta _0}
 \end{equation}
where $f ({\bf R})$ is the node nonlinearity and $D_ff({\bf R})$ is the derivative of $f$ with respect to the node variables. 

\begin{figure}
\centering
\includegraphics[scale=0.8]{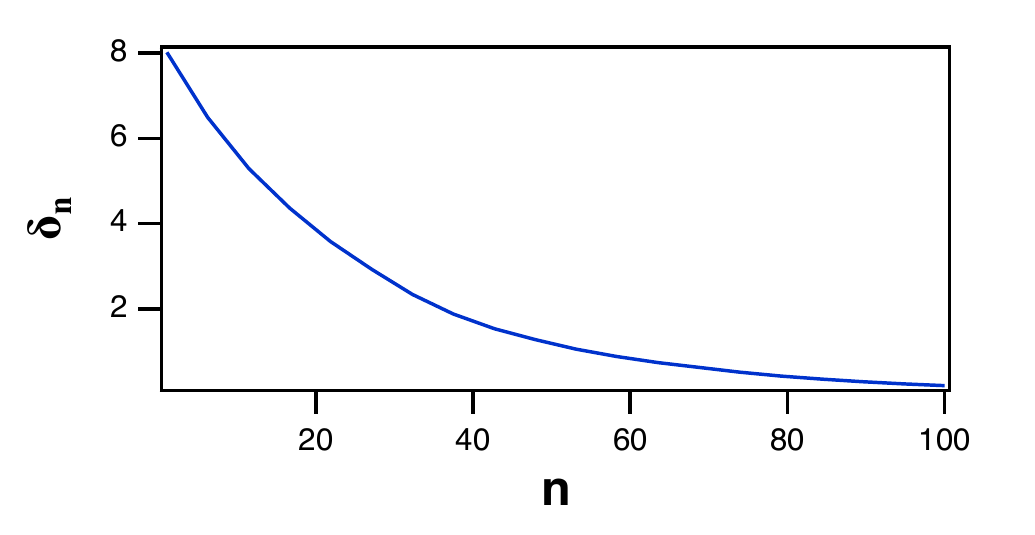} 
  \caption{ \label{perturb} Norm of the perturbation to the tanh reservoir computer as a function of number of time steps $n$. For this plot, $\varepsilon=1$ and $g=1$.}
  \end{figure}

To create a statistic based on the variational equation, for a system with $M$ nodes I created a random $M \times M$ matrix for $\delta_0$ and then made the rows orthonormal. I propagated this matrix with the nonlinear variational equation. {I propagate the variation along a known trajectory of the reservoir computer and label the initial variation with $j$ to indicate the particular initial condition. I then calculated the norm of $\delta_n(j)$ after $n$ steps. The nonlinear variational equation depends on the reservoir variables, so I randomly picked 100 different starting points on the reservoir attractor ($j = 1, \ldots, 100$) and averaged the resulting norm. }In order to make the statistic similar to the memory capacity statistic, I chose a maximum delay $\tau_{max}$, which I set equal to 100 to match the delay for the memory capacity calculation. A typical perturbation as a function of number of time steps $n$ for the tanh reservoir computer is plotted in figure \ref{perturb}.
I then take the sums
\begin{equation}
\label{varsum}
 \begin{array}{l}
{\delta_{{\mathop{\rm sum}} } = \sum\limits^{{\tau _{\max }}}_{n=1} {\left\| {\delta _n(j)} \right\|} }
\end{array}
 \end{equation}

where the $\left\| \; \right\|$ operator returns the 2-norm of a matrix. The norm of the variation, $D_{var}$, is the mean of the individual sums
\begin{equation}
\label{varnorm}
{D_{var}=\frac{1}{100} \sum\limits_{j=1}^{100 }\delta_{{\mathop{\rm sum}}} (j)}
\end{equation}

While this statistic, which I call the norm of the variation, is similar to the largest Lyapunov exponent, it does not follow exactly the same pattern. At short delays, the decay of the variation will be dominated by the most negative Lyapunov exponents, while the less negative exponents will govern the decay at later times.

While the norm of the variation does not depend on signal fitting and uses the actual signal driving the reservoir, it can only be found if one has the equations defining the reservoir. The delay capacity method described in the next section can be used when only experimental data is available.

\subsection{Delay Capacity}
The delay capacity statistic is adapted from the consistency capacity developed in \cite{jungling2021}. The consistency capacity method used the auxiliary system approach to detecting generalized synchronization \cite{Abarbanel:1996} to estimate the different computational capacities of a reservoir computer due to different input signals. J{\"u}ngling et al. \cite{jungling2021} apply a whitening transformation to the matrix ${\bf R}$ of signals from the reservoir computer. Typically they create two copies of the reservoir and use two different input signals, $s_1(n)=s(n)+\eta$ or $s_2(t)=s(n)+\eta '$, where $\eta$ and $\eta'$ are different noise signals. They find the covariance matrices for the two reservoirs, add a small regularization constant and then use the covariance matrices to create whitening transforms for the two reservoirs. After whitening, they find the cross covariance between the whitened reservoirs and use the trace of the cross covariance as a measure of capacity.

To estimate the delay capacity, rather than create two copies of a reservoir computer, I compared the reservoir computer signals at two different times. I whiten both of these sets of signals, calculate the cross covariance and find the trace. Again, to be consistent with the definition of memory capacity I sum the capacities for all delays and divide by the number of delays.

The signals from a reservoir with $M$ nodes and $N$ time series points may be arranged in an $M \times N$ matrix ${\bf R}_0$
\begin{equation}
\label{omega}
{{\bf R} _0} = \left[ {\begin{array}{*{20}{c}}
{{r_1}\left( 1 \right) - {{\bar r}_1}}& \cdots &{{r_1}\left( N \right) - {{\bar r}_1}}\\
{{r_2}\left( 1 \right) - {{\bar r}_2}}&{}&{{r_2}\left( N \right) - {{\bar r}_2}}\\
 \vdots &{}& \vdots \\
{{r_M}\left( 1 \right) - {{\bar r}_M}}& \cdots &{{r_M}\left( N \right) - {{\bar r}_M}}
\end{array}} \right]
\end{equation}
where $\bar r_i$ is the mean of $r_i$. This matrix is similar to the matrix $\Omega$ in eq. (\ref{fitmat}) except that it does not contain any terms in $r_i^2$ because these do not contribute to the memory.

The covariance matrix is then formed as:
\begin{equation}
\label{cmat}
{\bf C} = \frac{{{{\bf R}_0(t) }{\bf R}_0^T(t) }}{N}+L_{reg}{\bf I} .
\end{equation}
The regularization factor $L_{reg}=10^{-10}$ is added because the covariance matrix can be near singular.

 The covariance matrix is then decomposed by a singular value decomposition, ${\bf C} = {\bf{US}}{{\bf{V}}^{\bf{T}}}$ and the reservoir signals are normalized as
\begin{equation}
\label{resnorm}
{\widetilde {\bf{R}}_0} ={\left( {{{\bf{V}}^T}{{\bf{R}}_0}} \right)}\left({\sqrt {\bf{S}} }\right)^{-1}
\end{equation}

As a measure of memory, a second matrix is created, ${{\bf R} _\tau }\left( t \right) = {{\bf R} _0}\left( {t - \tau } \right)$ and normalized in the same manner to produce ${\tilde {\bf R} _\tau }\left( t \right)$. The cross covariance between the normalized versions of the regular and delayed matrices is then found

\begin{equation}
\label{crosscorrmat}
{\bf{C}}\left( \tau  \right) = \frac{{{{\tilde {\bf R} }_0}\tilde {\bf R} _\tau ^T}}{N} .
\end{equation}

\begin{figure}
\centering
\includegraphics[scale=0.8]{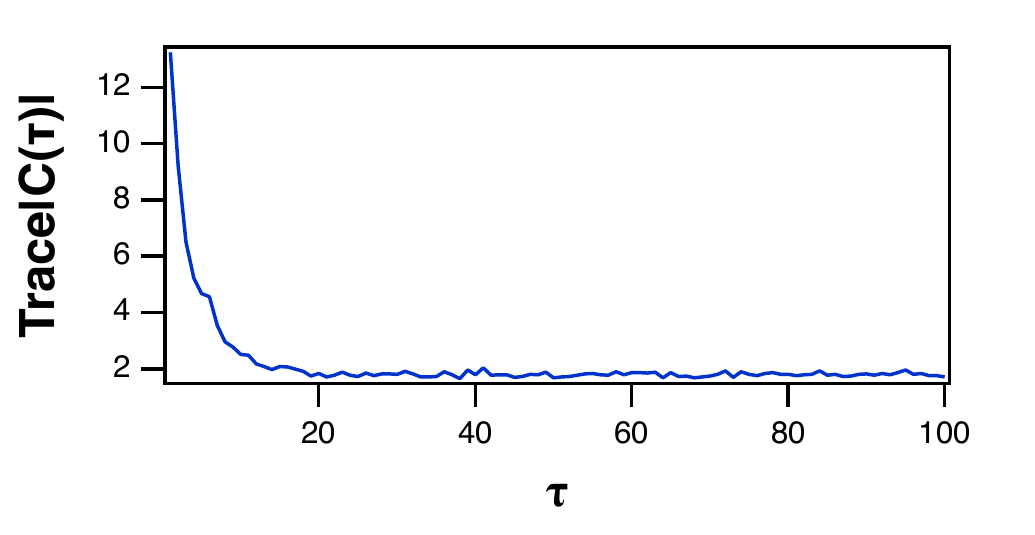} 
  \caption{ \label{cortrace}  Trace of the cross covariance matrix ${\bf C}(\tau)$ as a function of the delay $\tau$, for the tanh reservoir computer. In this plot, $\varepsilon=1$ and $g=1$.}
  \end{figure}

The delay capacity statistic is calculated as
\begin{equation}
\label{delcap}
{\Theta _d} = \frac{{\sum\limits_{\tau  = 0}^{{\tau _{\max }}} {{\rm{Trace}}\left| {{\bf{C}}\left( \tau  \right)} \right|} }}{{{\tau _{\max }}}}
\end{equation}
where the $\left| \; \right|$ operator indicates an absolute value. The trace of the cross covariance matrix ${\bf C}(\tau)$ is shown in figure \ref{cortrace}.

{
\subsection{Nonlinear Index}
The nonlinearity in a reservoir computer is important in reproducing signals from nonlinear systems, so as an additional statistic I used the nonlinear index described in \cite{dambre2012}. The reservoir was driven with sine waves with 100 different periods ranging from $T_j=10 , j=1$ to $T_j=50, j=100$. The individual node signals were Fourier transformed into ${\mathcal F}_i(f)$. For a fundamental frequency $f_j=2\pi T_j$, the nonlinear index is
\begin{equation}
\Gamma \left( {{f_j}} \right) = \frac{1}{M}\sum\limits_{i = 1}^M {\left( {\frac{{\sum\limits_{f > {f_j}} {\left| {{\mathcal F_i}(f)} \right|} }}{{\left| {{\mathcal F_i}({f_j})} \right|}}} \right)} 
\end{equation}
where the $\left| \; \right|$ operator returns the absolute magnitude. The mean nonlinear index is 
\begin{equation}
\label{nlindex}
\Gamma  = \frac{1}{{100}}\sum\limits_{j = 1}^{100} {\Gamma \left( {{f_j}} \right)} .
\end{equation}
}

\section{Simulations}

\subsection{Fitting the Lorenz System}
In order to see how the reservoir parameters affect the actual performance, the Lorenz  chaotic system was used to drive the reservoir.

\begin{figure*}
\centering
\includegraphics[scale=0.8]{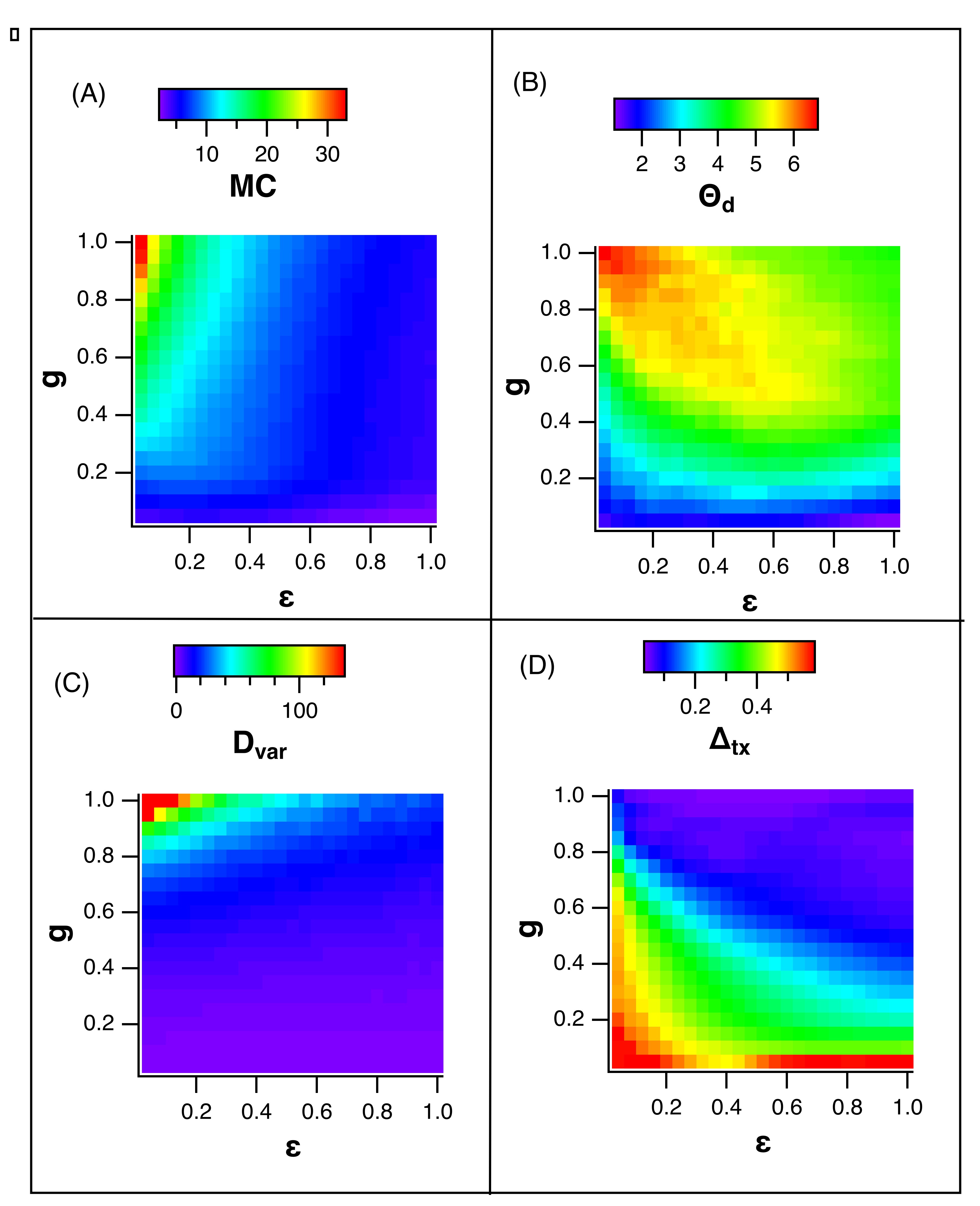} 
  \caption{ \label{lormemerr} {Testing error and memory statistics for the tanh reservoir driven by the Lorenz $x$ signal and trained on the Lorenz $z$ signal. (A) is the memory capacity MC, (B) is the delay capacity $\Theta_d$ as defined in eq. (\ref{delcap}) , (C) is the variation of the norm $D_{var}$ (eq (\ref{varnorm})) while (D) is the testing error $\Delta_{tx}$.}}
  \end{figure*}

The Lorenz system is described by \cite{lorenz1963}
\begin{equation}
\label{loreq}
\begin{array}{l}
\frac{{dx}}{{dt}} = {p_1}\left( {y - x} \right)\\
\frac{{dy}}{{dt}} = x\left( {{p_2} - z} \right) - y\\
\frac{{dy}}{{dt}} = xy - {p_3}z
\end{array}
\end{equation}
with $p_1=10$, $p_2=28$ and $p_3=8/3$. The equations were numerically integrated with a time step of 0.02. The input to the reservoir computer was $x$, and the reservoir computer was trained on $z$. The testing error for fitting the Lorenz $z$ signal is plotted in figure \ref{lormemerr}.

{
The three different memory statistics or the Lorenz system are also shown in figure \ref{lormemerr}. }The delay capacity $\Theta_d$ in figure \ref{lormemerr} roughly echos the memory capacity MC , but because the delay capacity depends on the actual input signal, it is not the same as the memory capacity. The delay capacity is a measure of how long slowly the autocorrelation of signals in a reservoir computer drops off with time, while memory capacity is measured by how well the reservoir computer fits a delayed noise signal. The delay capacity is affected by the autocorrelation of the input signal. 

The variation of the norm in figure \ref{lormemerr} does more closely resemble the plot of memory capacity. The variation of the norm measures how quickly a perturbation to the reservoir trajectory decays, but unlike the memory capacity, it depends on the signal that drives the reservoir. The decay of a perturbation would seem to be an excellent way to quantify fading memory, but calculating the norm of the variation requires that the reservoir computer equations are known, which may not be true in an experiment.

\begin{figure}
\centering
\includegraphics[scale=0.8]{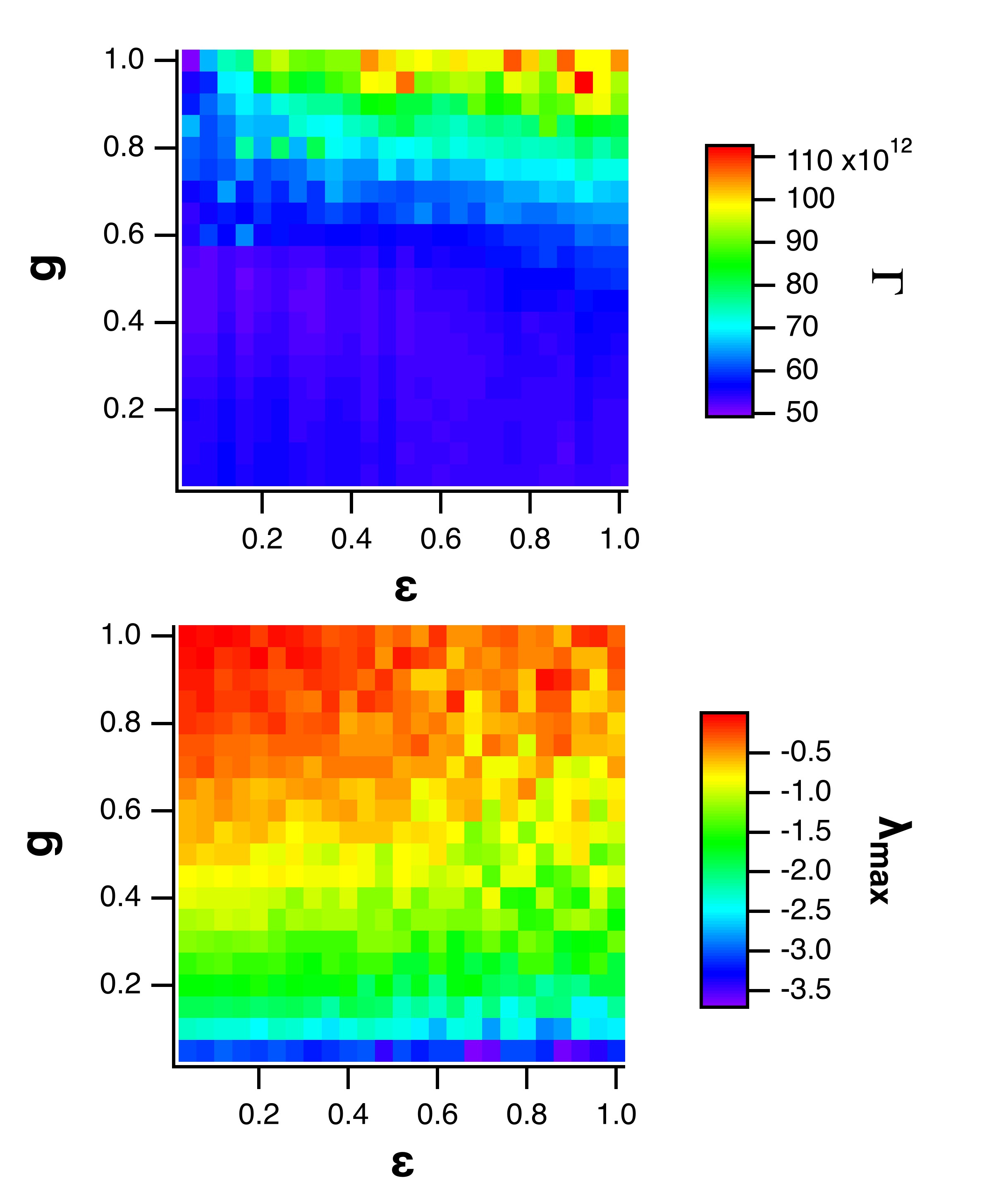} 
  \caption{ \label{lornonlinly} {The top plot is the nonlinear index $\Gamma$ for the reservoir computer driven by the Lorenz $x$ signal, while the bottom plot is the largest Lyapunov exponent. }  }
  \end{figure}

The testing error for the Lorenz system is not minimized at the largest values of the memory capacity, shown in figure \ref{lormemerr}. The different memory statistics in figure \ref{lormemerr} are all maximized for large values of the feedback constant $g$ but small values of the input multiplier $\varepsilon$.
 The minimum training error comes at larger values of $\varepsilon$ than the memory capacity. A reason for this may be seen in figure \ref{lornonlinly}, which shows the nonlinearity index $\Gamma$ and the largest Lyapunov exponent for the reservoir computer driven by the Lorenz $x$ signal.

 {Figure \ref{lornonlinly} shows that the nonlinear index $\Gamma$ increases as the input multiplier $\varepsilon$ increases. For very small signals the tanh nonlinearity is approximately linear, while fitting the Lorenz $z$ signal requires some nonlinearity, so increasing the nonlinearity in the reservoir computer by increasing the input multiplier decreases the testing error. This is one reason why the smallest testing error in figure \ref{lormemerr} does not come for the largest memory capacity. Some work such as \cite{inubushi2017,verstraeten2010} has suggested that increasing nonlinearity can decrease memory capacity. }

The largest Lyapunov exponent for the Lorenz driven tanh reservoir computer, also in figure \ref{lornonlinly}, is calculated from the variational equation as is the norm of the variation, but the two statistics follow different patterns. For a reservoir computer with $M$ nodes, there are $M$ Lyapunov exponents. The initial decay of a perturbation may be governed by the most negative of these exponents; only after the degrees of freedom governed by the most negative exponents have decayed away does the largest exponent dominate. The norm of the variation drops off faster than what would be expected from the largest Lyapunov exponent because of the more negative exponents, but as expected, the largest value of all the memory statistics comes when the largest reservoir Lyapunov exponent is the least negative.

 \section{Increasing Memory}
\label{moremem} 
 One way to increase the memory capacity of a reservoir computer is to operate near the edge of stability, where the largest Lyapunov exponent for the reservoir computer is just below 0. Working too close to the edge of stability can create problems, however.  In \cite{carroll2020b}, it was demonstrated that as the reservoir computer Lyapunov exponents increased, they overlapped with the Lyapunov exponent spectrum of the training system, increasing the fractal dimension of the reservoir signals and causing a larger training error. In a different approach, in \cite{farkas2016}  the memory capacity was increased by making the columns in the adjacency matrix orthogonal. In \cite{verstraeten2007} the memory capacity of different types of nodes were compared as the reservoir computer parameters were varied.
 
\subsection{Sparsity}
Increasing the sparsity of the adjacency matrix; that is, decreasing the number of connections between nodes- could possibly increase the memory of the reservoir computer. Conventional rules for designing a reservoir computer often call for using a sparse adjacency matrix, although no justification is usually given for this rule. Using an adjacency matrix with fewer connections means that the feedback path between nodes will be longer, which may increase memory.

\begin{figure}
\centering
\includegraphics[scale=0.8]{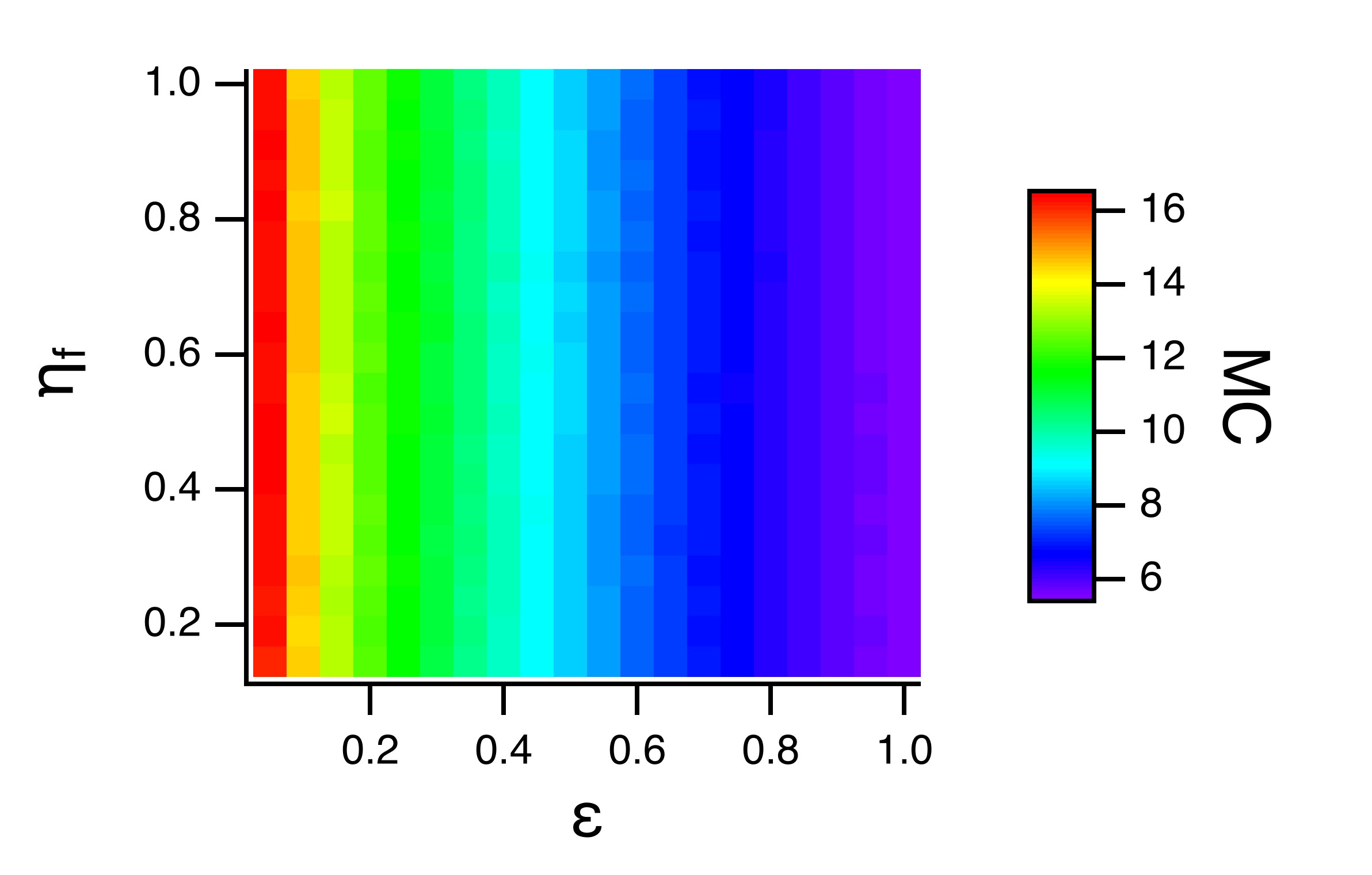} 
  \caption{ \label{memcapsparse} The memory capacity for the tanh reservoir of eq. (\ref{tanhnode}) as the fraction of nonzero entries in the adjacency matrix. $\eta_f$, varies.    }
  \end{figure}

\begin{figure*}
\centering
\includegraphics[scale=0.8]{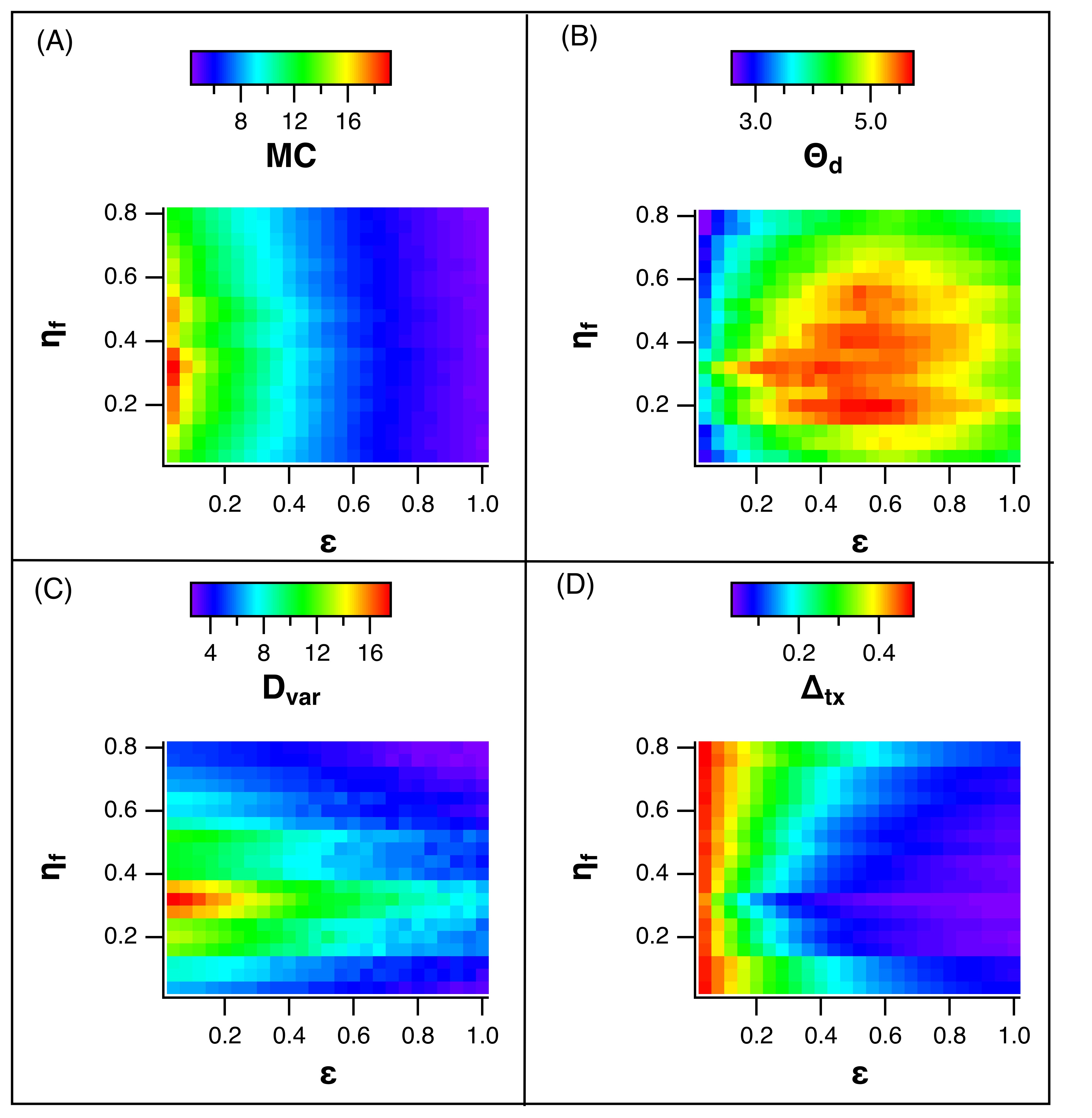} 
  \caption{ \label{lorfixlen} {(A) The memory capacity for the tanh reservoir computer of eq. (\ref{tanhnode}), driven by the Lorenz $x$ signal and trained on the Lorenz $z$ signal, as the fraction of nonzero entries in the adjacency matrix. $\eta_f$, varies. (B) is the delay capacity, (C) is the norm of the variation, while (D) is the testing error. The mean weighted path length $<L_W>$ for these plots was fixed at 2.0. } }
  \end{figure*}

Farka{\v s} et al. \cite{farkas2016} examined adjacency matrices with different amounts of sparsity, but found that changing the sparsity did not change the maximum memory capacity. While figure \ref{memcapsparse} below agrees with this conclusion, I show that changing the sparsity of the adjacency matrix also alters the strength of the interaction between nodes. If the strength of this interaction is compensated for, then the effect of sparsity on the memory capacity can be seen in figure \ref{lorfixlen}. 

A series of adjacency matrices were created for the tanh reservoir of eq. (\ref{tanhnode}) by randomly choosing a fraction $\eta_f$ of the connections between nodes and setting them to a number drawn from a Gaussian random distribution with mean 0 and standard deviation 1. For the smallest values of $\eta_f$ it was possible for some nodes to have no connections to any other nodes, so the adjacency matrix was required to have at least one entry in each row and at least one entry in each column.

The path length between nodes should depend on $\eta_f$. The length of the path between two nodes in a network is a well known statistic \cite{albert2002}, and may be found by a breadth-first search of the adjacency matrix \cite{Skiena2008}. The unweighted path length between any nodes i and j is designated as $L_U(i,j)$.  For node $i_0$, the following algorithm puts the unweighted path length to all the other nodes into the vector distanceList:

\begin{algorithmic}
\STATE distanceList($1 \ldots M$) $\gets \infty$
\STATE distanceList$(i_0)\gets 0$
\STATE queue $\gets i_0$

\WHILE{queue is not empty}
	\STATE queue2 is empty
	\FOR{$k=1$ to length of queue}
		\STATE  $i_k \gets$ queue($k$)
		\STATE inList $\gets A_{i_k,1 \ldots M}$
		\STATE outList $\gets A_{1 \ldots M, i_k}$
		\STATE nodeList $\gets$ inList $\cup$ outList
		\FOR {$j=1$ to length of nodeList}
			\STATE $i_M \gets$ nodeList($j$)
			\IF{distanceList($i_M$) $\ne \infty$}
				\STATE add $i_M$ to queue2
				\STATE distanceList($i_M$) $\gets$ distanceList($i_k$) $+ 1$
		
			\ENDIF
		\ENDFOR
	\ENDFOR
	\STATE queue $\gets$ queue2
\ENDWHILE
\end{algorithmic}

The mean unweighted path length is
\begin{equation}
\label{unwpath}
\left\langle {{L_U}} \right\rangle  = \frac{1}{M}\sum\limits_{i = 1}^M {{\rm{distanceList}}\left( i \right)} 
\end{equation}

\begin{figure}
\centering
\includegraphics[scale=0.8]{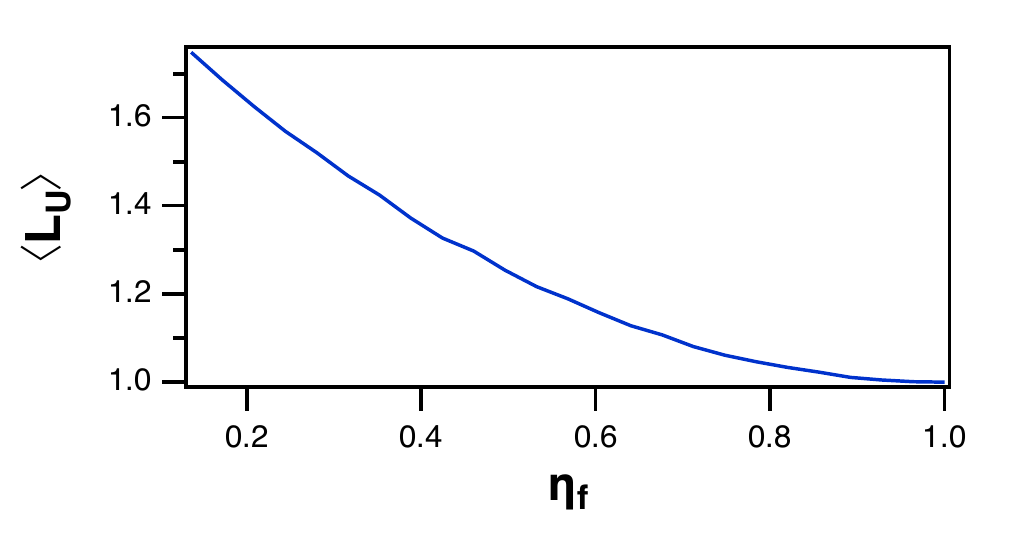} 
  \caption{ \label{unwpathfig} Mean unweighted path length $<L_U>$ for adjacency matrices versus the fraction $\eta_f$ of the entries that are nonzero. }
  \end{figure}

If there is no path between two nodes, the path length is  $\infty $ and it is not used in calculating the mean path length. Distances with values of 0 are also not included in the mean unweighted path length. Figure \ref{unwpathfig} shows the mean unweighted path length $<L_U>$ as a function of $\eta_f$.

Figure \ref{memcapsparse} shows the memory capacity for the tanh reservoir as the fraction $\eta_f$ varies. The amplitude parameter $g$ was set to 1.0.  

  The memory capacity in figure \ref{memcapsparse} is not correlated with the unweighted path length in figure \ref{unwpathfig}. As the number of connections between nodes changes, the strength of the interaction between nodes also changes, so just changing the sparsity of the adjacency matrix does not change the memory capacity.

 In  \cite{carroll2020a} a weighted path length statistic was derived that was a measure of the strength of the interaction between nodes. The weighted path length statistic $L_W(i,j)$ was calculated in a similar fashion to the unweighted path length statistic, except that the distance between two adjacent nodes is not 1 but
 \begin{equation}
\label{dist2}
{\delta _{ij}} = \ln \left[ {\frac{{1}}{{\left| {{A_{ij}}} \right| + \left| {{A_{ji}}} \right|}}} \right].
\end{equation}

The distance was defined in this way so that the weighted distance between nodes that were more strongly coupled was smaller. Because of the natural log in eq. (\ref{dist2}) the weighted path length can have negative values, so it is not a true distance. The log factor was included because the weighted path length can vary over several orders of magnitude.

\begin{figure}
\centering
\includegraphics[scale=0.8]{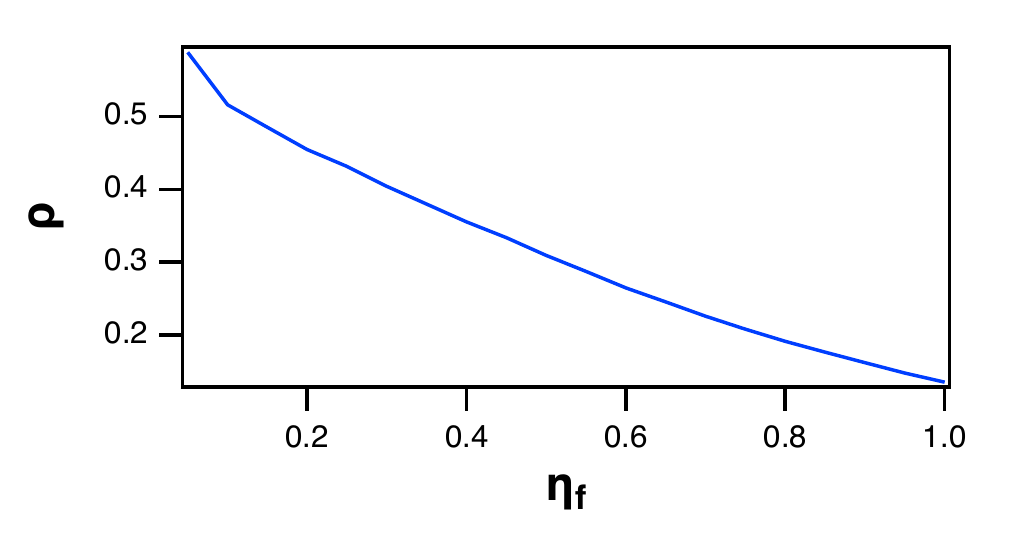} 
  \caption{ \label{spectrad} The spectral radius for the tanh reservoir of eq. (\ref{tanhnode}) as the fraction of nonzero entries in the adjacency matrix. $\eta_f$, varies. The spectral radius was varied to keep the  mean weighted path length $<L_W>$ fixed at 2.0.  }
  \end{figure}

The interaction between nodes may be held constant by varying the spectral radius so that the mean weighted path length $<L_W>$ is constant. The resulting spectral radius $\rho$ is shown in figure \ref{spectrad}. Figure \ref{lorfixlen} shows the memory capacity when the spectral radius for the adjacency matrix was varied to fix the mean weighted path length $<L_W>$ at 2.0. This value was chosen because it was the mean weighted path length when $\eta_f=0.5$. 

 When the mean weighted path length $<L_W>$ was fixed, the memory capacity for the tanh reservoir showed a variation with the node fraction $\eta_f$, but the relationship was not monotonic. Figure \ref{lorfixlen} shows that for small values of the input multiplier $\varepsilon$ the memory capacity reaches a maximum for $\eta_f$ near 0.3. The maximum is less prominent as $\varepsilon$ increases. The other measures of memory plotted in figure \ref{lorfixlen} also had maxima near $\eta_f=0.3$. In \cite{farkas2016} a similar pattern was seen; the memory capacity went through a maximum as the spectral radius was varied. The testing error, by contrast, was a maximum at the largest values of the input constant $\varepsilon$. It was seen in figure \ref{lornonlinly} that larger values of nonlinearity led to lower testing error for the Lorenz system.
 
 \subsection{R{\" o}ssler System}
 The effect of the reservoir computer memory depends on the problem being solved; as a demonstration, the same reservoir computer was driven with the $x$ signal from the Rossler chaotic system and trained on the $z$ signal. The R{\"o}ssler system is described by \cite{rossler1976}
\begin{equation}
\label{rosseq}
\begin{array}{*{20}{l}}
{\frac{{dx}}{{dt}} =  - y - {p_1}z}\\
{\frac{{dy}}{{dt}} = x + {p_2}y}\\
{\frac{{dz}}{{dt}} = {p_3} + z\left( {x - {p_4}} \right)}
\end{array}
\end{equation}

These equations were numerically integrated  with a time step $t_s$=0.3, and parameters $p_1=1$, $p_2=0.2$, $p_3=0.2$, $p_4=5.7$.
 
 \begin{figure*}
\centering
\includegraphics[scale=0.8]{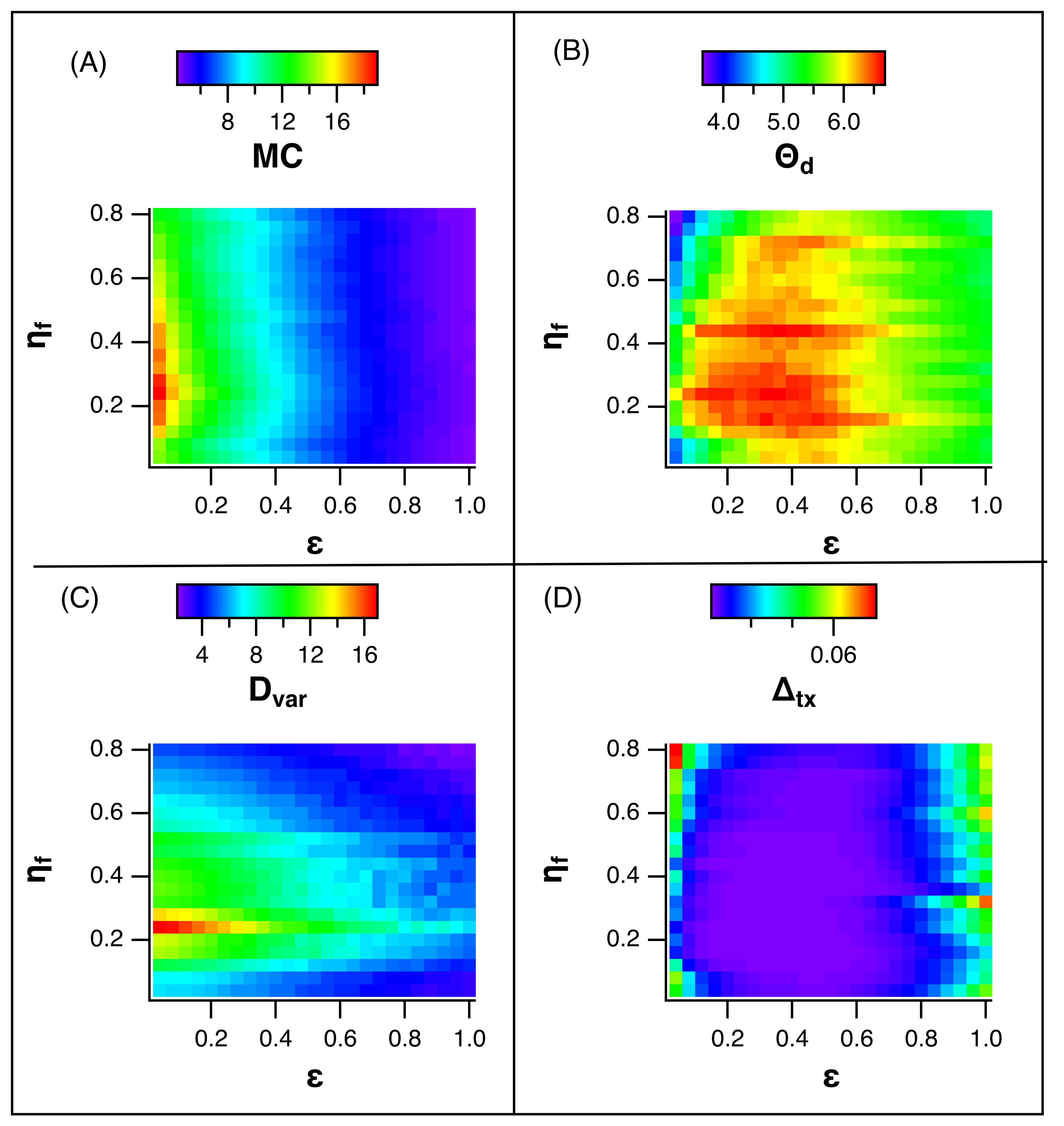} 
  \caption{ \label{rossfixlen} {(A) The memory capacity for the tanh reservoir computer of eq. (\ref{tanhnode}), driven by the R{\"o}ssler $x$ signal and trained on the R{\"o}ssler $z$ signal, as the fraction of nonzero entries in the adjacency matrix. $\eta_f$, varies. (B) is the delay capacity, (C) is the norm of the variation, while (D) is the testing error. The mean weighted path length $<L_W>$ for these plots was fixed at 2.0. } }
  \end{figure*} 

\begin{figure}
\centering
\includegraphics[scale=0.8]{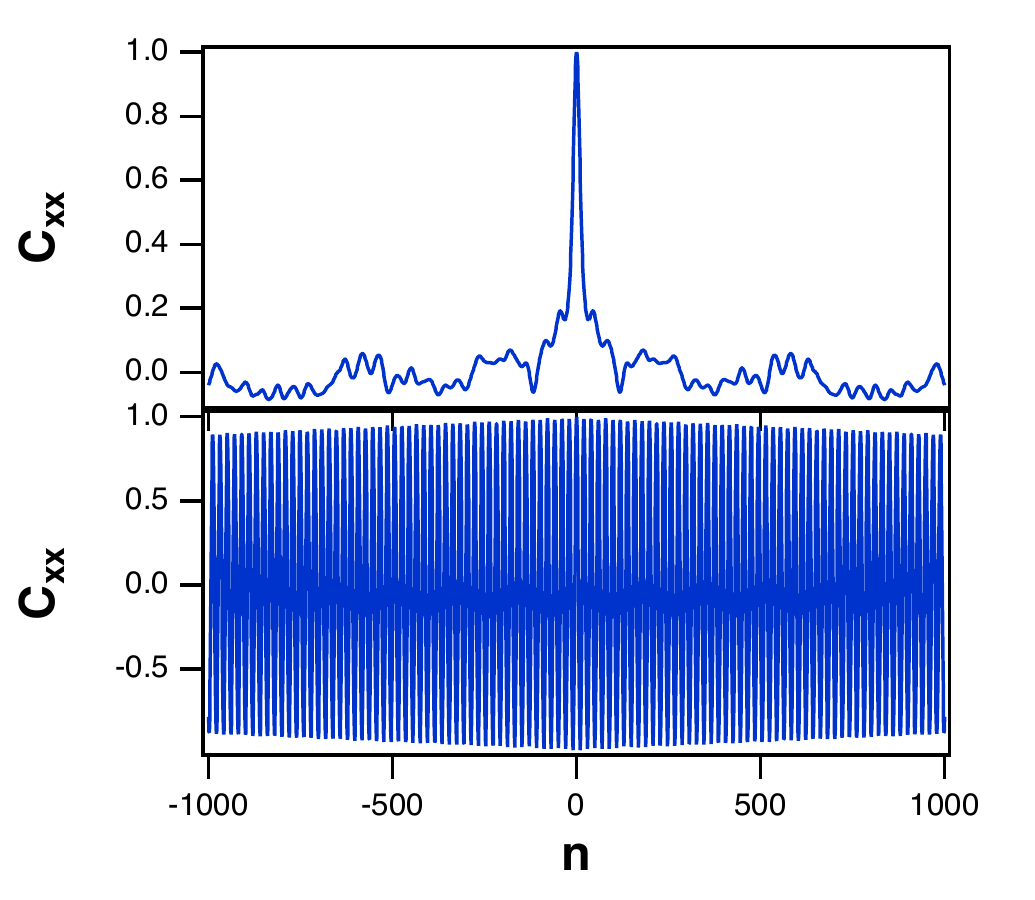} 
  \caption{ \label{xcorr} Top plot is the autocorrelation of the Lorenz $x$ signal, $C_{xx}$, while the bottom plot is the autocorrelation of the R{\"o}ssler $x$ signal. The horizontal axis is labeled in the number of time series points $n$. For the top plot, each time step was 0.02, while in the bottom plot is was 0.3. }
  \end{figure}

The pattern of memory as a function of $\varepsilon$ and $\eta_f$ when the reservoir computer is driven with a R{\"o}ssler $x$ signal is similar to when it was driven with a Lorenz $x$ signal, but the pattern of testing error is very different. For the R{\"o}ssler system the minimum training error does come when the memory is maximized.   A possible reason the patterns of testing error in the Lorenz and R{\" o}ssler systems are so different is because of their respective autocorrelation functions. The top plot in figure \ref{xcorr} shows the autocorrelation for the Lorenz $x$ signal, while the bottom plot shows the autocorrelation for the R{\"o}ssler $x$ signal.

Figure \ref{xcorr} shows that the Lorenz autocorrelation drops quickly to near 0, while for the R{\"o}ssler the autocorrelation is large for large times. The  R{\"o}ssler signal is almost periodic, so the testing error will be small when the reservoir signals maintain a larger correlation over time. For the Lorenz signal, with its autocorrelation that quickly decreases, a reservoir computer with a long memory will mix together parts of the signal that are uncorrelated in time, resulting in larger errors. In \cite{carroll2020b} a similar effect was attributed to an overlap in the Lorenz exponent spectrum; these are two different ways to explain the same thing. For the R{\"o}ssler system, whose autocorrelation persists over long times, more memory leads to lower error.
 
 The different measures of memory in figures \ref{lorfixlen} and \ref{rossfixlen} show that the memory capacity goes through a maximum when the fraction of occupied nodes is about 0.3. The next section provides a possible cause of this maximum.
 
 \subsection{Approximate Delay Coefficients}
 \label{approxcoeff}
 The reason for this maximum in the memory capacity may be shown with a linear model of a reservoir computer. Consider a simple linear reservoir computer:
 \begin{equation}
 \label{linres}
{\bf{R}}\left( {n + 1} \right) = \rho {\bf{AR}}\left( n \right) + {\bf{W}}s\left( n \right)
 \end{equation}
 where ${\bf R}$ is the vector of reservoir variables, ${\bf A}$ is the adjacency matrix, $\rho$ is used to rescale the spectral radius of ${\bf A}$, ${\bf W}$ is the input vector and the input signal is $s(n)$. The input vector ${\bf W}$ was set to all ones. The adjacency matrix in eq. (\ref{linres}) was rescaled so that the absolute magnitude of its largest eigenvalue was 1.0, so the value of $\rho$ will set the desired spectral radius.
 
If the reservoir computer output is truncated after five iterations with ${\bf R}(0)=0$, it looks like
\begin{equation}
\label{liniter}
\begin{array}{*{20}{l}}
{{\bf{R}}\left( {n + 1} \right) = {\bf{W}}s\left( n \right) + {{\bf{b}}_1}s(n - 1) + {{\bf{b}}_2}s\left( {n - 2} \right) + {{\bf{b}}_3}s\left( {n - 3} \right) + {{\bf{b}}_4}s\left( {n - 4} \right)}\\
{{{\bf{b}}_1} = {\rho ^{}}{\bf{AW}}}\\
{{{\bf{b}}_j} = {\rho ^{j - 1}}{{\bf{A}}^{j - 1}}{{\bf{b}}_1}\quad j > 1}
\end{array}
\end{equation}

\begin{figure}
\centering
\includegraphics[scale=0.8]{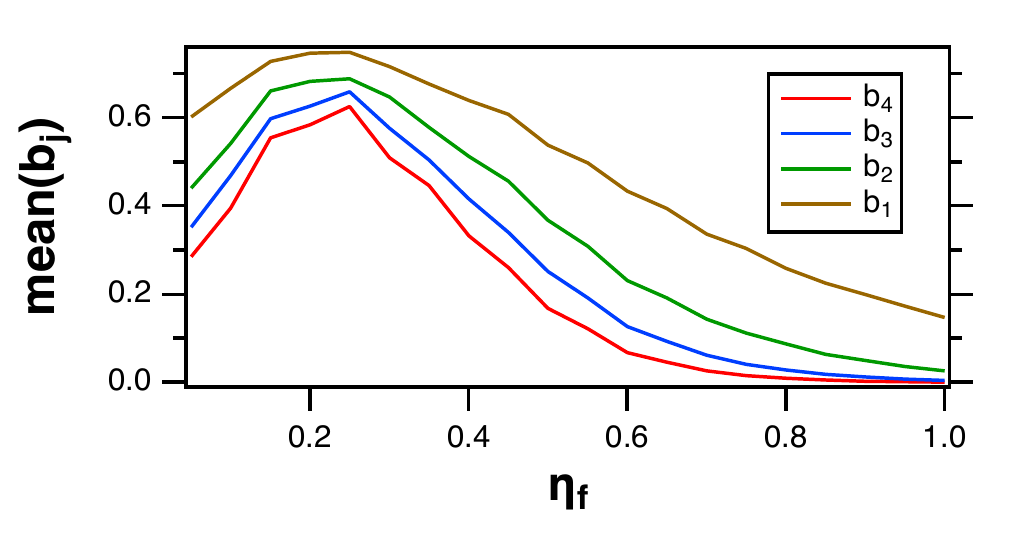} 
  \caption{ \label{delcoeff} Values of the delay coefficients ${\bf b}_j$ from eq. (\ref{liniter}), averaged over all nodes.  }
  \end{figure}

The delay coefficients from eq. (\ref{liniter}) were calculated by varying the fraction of occupied nodes $\eta_f$ and substituting the spectral radius from figure \ref{spectrad} for $\rho$. The coefficients for the delayed input signals $s(n-1)$ to $s(n-4)$, averaged over all nodes,  are shown in figure \ref{delcoeff}.

  The delay coefficients all peak for $\eta_f$ near 0.3. The coefficient on the undelayed signal $s(n)$ is 1.0, so the memory capacity is partly determined by the ratio of the delayed signal to the undelayed signal. The approximation of eqs. (\ref{linres}-\ref{liniter}) does not exactly match the behavior of the tanh reservoir because the nonlinearity will also affect the memory capacity, but the linear approximation can help explain why the memory capacity goes through a maximum.

  \section{Multidimensional Nodes}
    \label{mdtanh}
 Changing the sparsity of the adjacency matrix can change the memory capacity of a reservoir computer, but only to a limited extent. In a nonlinear system, changing the sparsity causes changes in other quantities, such as the mean weighted path length between nodes. A different way to alter the memory capacity is to add extra dynamical dimensions to the nodes. 
  
  Multidimensional nodes may be created by adding delayed coordinates to the tanh nodes from eq. (\ref{tanhnode}):
 \begin{equation}
 \label{ndtanh}
 \begin{array}{l}
{r_{i,1}}\left( {n + 1} \right) = g\tanh \left( {\sum\limits_{j = 1}^M {{A_{i,j}}{r_{i,1}}\left( n \right)}  + \varepsilon s\left( n \right) + 0.5{r_{i,{d_e}}}} \right)\\
{r_{i,j}}(n + 1) = {r_{i,j - 1}}(n)\quad j = 2 \ldots {d_e}
\end{array}
 \end{equation}
 
 where $d_e$ is the dimension of the node activation function and $r_{i,j}(n)$ is the $j$th component of the $i$'th node. One could create similar nodes for an ordinary differential equation system by using a delayed signal.
 
\begin{figure}
\centering
\includegraphics[scale=0.8]{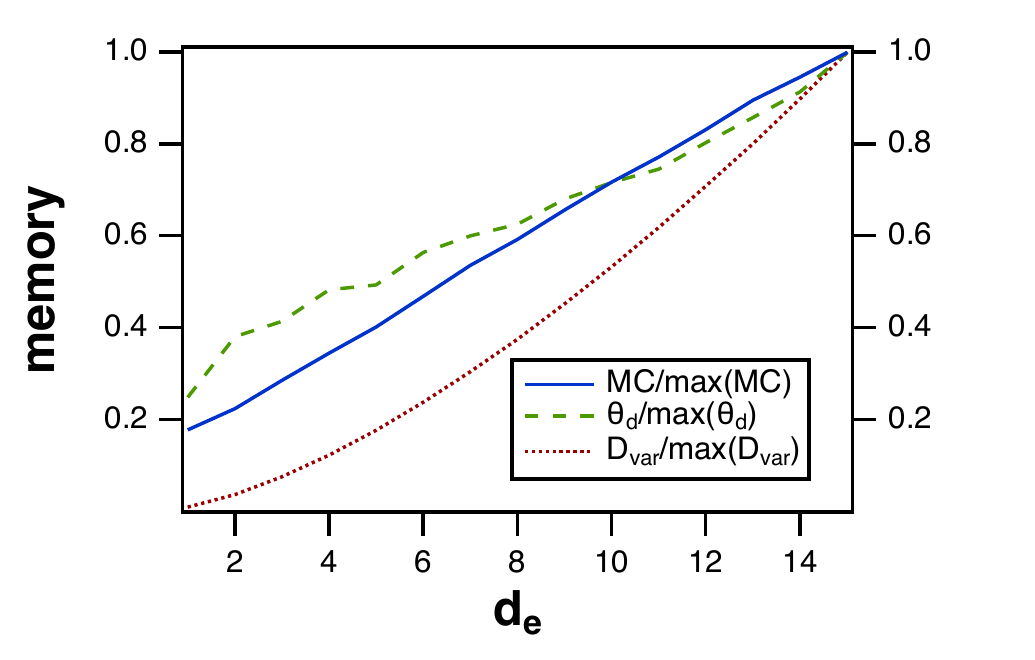} 
  \caption{ \label{tanhvardmem} {The memory capacity MC, the delay capacity $\theta_d$ and the norm of the variation $D_{var}$ for the reservoir computer with multidimensional nodes. All statistics were normalized by their maximum values so the could be plotted on one axis. The input signal was the Lorenz $x$ signal.}}
  \end{figure}

 Figure \ref{tanhvardmem}  shows the three different measures of memory used in this work for the multidimensional tanh nodes of eq. (\ref{ndtanh}) as the node dimension $d_e$ was scanned and the parameter $g$ was fixed at 0.35, $\varepsilon=0.5$ and the spectral radius $\rho=1$. {These parameter values were set by scanning through a number of parameter values and node dimensions and choosing the values of $g$ and $\varepsilon$ that gave a minimum testing error for a range of $d_e$ between 5 and 10. }The input signal for all three statistics was a Lorenz $x$ signal.  To make it possible to plot all three statistics on one axis, all statistics were normalized by their maximum values. The memory variations were similar when the input signal was the R{\" o}ssler $x$ signal.
  
 All three memory statistics in figure \ref{tanhvardmem} are consistent, and all three show that increasing the node dimension $d_e$ increased the memory for the reservoir computer.
 
\subsection{Lorenz and R{\"o}ssler systems}
\label{ndnode}
\begin{figure}
\centering
\includegraphics[scale=0.8]{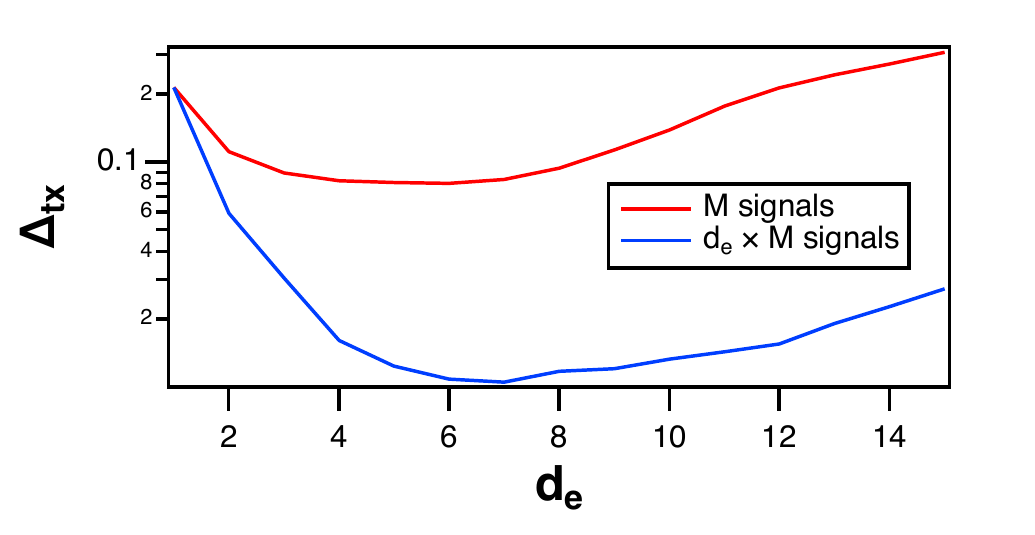} 
  \caption{ \label{lorerrvard} Testing error $\Delta_{tx}$ when the multidimensional reservoir computer is driven by the Lorenz $x$ signal and trained on the Lorenz $z$ signal. The dimension of the nodes in the reservoir computer was $d_e$. For the curve labeled $d_e$ signals, only the first component from each node, or $r_{i,1}(n)$, was used in the fit, while for the curve labeled  $d_e \times M$ signals, all $d_e$ components from each node were used.}
  \end{figure}

Figure \ref{lorerrvard} shows the testing error for the Lorenz system as the dimension $d_e$ of the reservoir computer was increased. There are two curves in figure \ref{lorerrvard}: the full reservoir computer produced $d_e \times M$ signals, so one curve shows the testing error found by using the full set of signals. Using more signals in testing the reservoir computer can by itself decrease the error, so to remove this effect, for the other curve only the signals corresponding to the first component from each node, $r_{i,1}(n)$, were used in the fit.

When only the first component from each node is used, the minimum testing error for the Lorenz system came for $d_e=6$, while when all components were used minimum was at $d_e=7$. The vertical axis on figure \ref{tanhvardmem} confirms that the different measures of memory all increase as the reservoir computer dimension $d_e$ increases, so there is an optimal amount of memory for fitting the Lorenz $x$ signal- more or less memory produces a larger training error.

\begin{figure}
\centering
\includegraphics[scale=0.8]{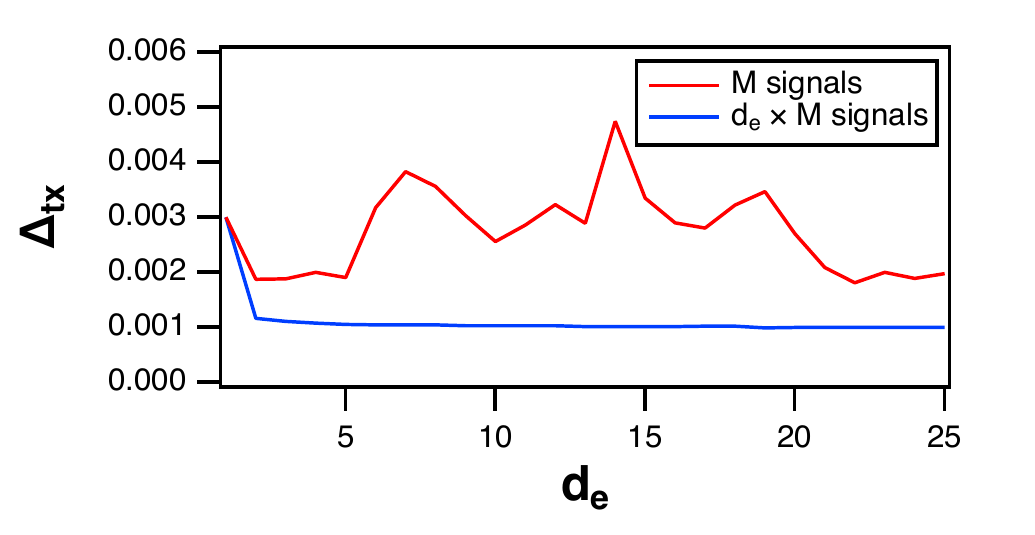} 
  \caption{ \label{rosserrvard} Testing error $\Delta_{tx}$ when the multidimensional reservoir computer is driven by the R{\"o}ssler $x$ signal and trained on the R{\"o}ssler $z$ signal. The dimension of the nodes in the reservoir computer was $d_e$. For the curve labeled $d_e$ signals, only the first component from each node, or $r_{i,1}(n)$, was used in the fit, while for the curve labeled  $d_e \times M$ signals, all $d_e$ components from each node were used.}
  \end{figure}

Figure \ref{rosserrvard} shows the testing error for the multidimensional reservoir computer when driven with the R{\"o}ssler system, as a function of the reservoir computer dimension, for both types of fits: when all $d_e \times M$ signals are used and when only the $M$ signals from $r_{i,1}$ are used.
The testing error for the R{\"o}ssler system does not go through a minimum, as did the testing error for the Lorenz system. The testing error for the R{\"o}ssler system when all $d_e \times M$ signals are used barely decreases as the reservoir computer dimension increases, while when only the first component for each node is used, there is scatter within the data but no decreasing trend. Increasing memory is not harmful in the R{\"o}ssler system, but there appears to be no significant improvement in the testing error from increasing memory.

The autocorrelations for the Lorenz and R{\"o}ssler systems were shown in figure \ref{xcorr}. The autocorrelation for the Lorenz system drops off quickly to near zero. If the reservoir computer memory is too long when attempting to fit the Lorenz signal, the reservoir computer will mix together parts of the Lorenz signal that are uncorrelated in time, degrading the testing error; as a result, there is an optimum memory capacity for fitting the Lorenz system. In contrast, the R{\"o}ssler system is nearly periodic, so its autocorrelation drops to zero only over a very long time. 
  
  \subsection{NARMA System}
\label{narmadat}
 The nonlinear autoregressive moving average (NARMA) system was developed as a way to test the ability of neuromorphic systems to reproduce a signal with memory \cite{atiya2000}. As used here the NARMA system is described by
\begin{equation}
\label{narmavar}
y\left( {n + 1} \right) = 0.3y\left( n \right) + 0.05y\left( n \right)\sum\limits_{j = 1}^{{N_N}} {y\left( {n - j} \right)}  + 1.5u\left( {n - {N_N} + 1} \right)u\left( n \right) + 0.1
\end{equation}
where the order of the model is $N_N$.   The input signal $u(n)$ is drawn from a uniform random distribution between 0 and 0.5.

 The dependance of the training error on both memory capacity and the memory required to reproduce a particular signal may be investigated by using NARMA  systems of varying orders. Equation \ref{narmavar} described a NARMA system of order $N_N$; in this section, $N_N$ varies from 1 to 10.

\begin{figure}
\centering
\includegraphics[scale=0.8]{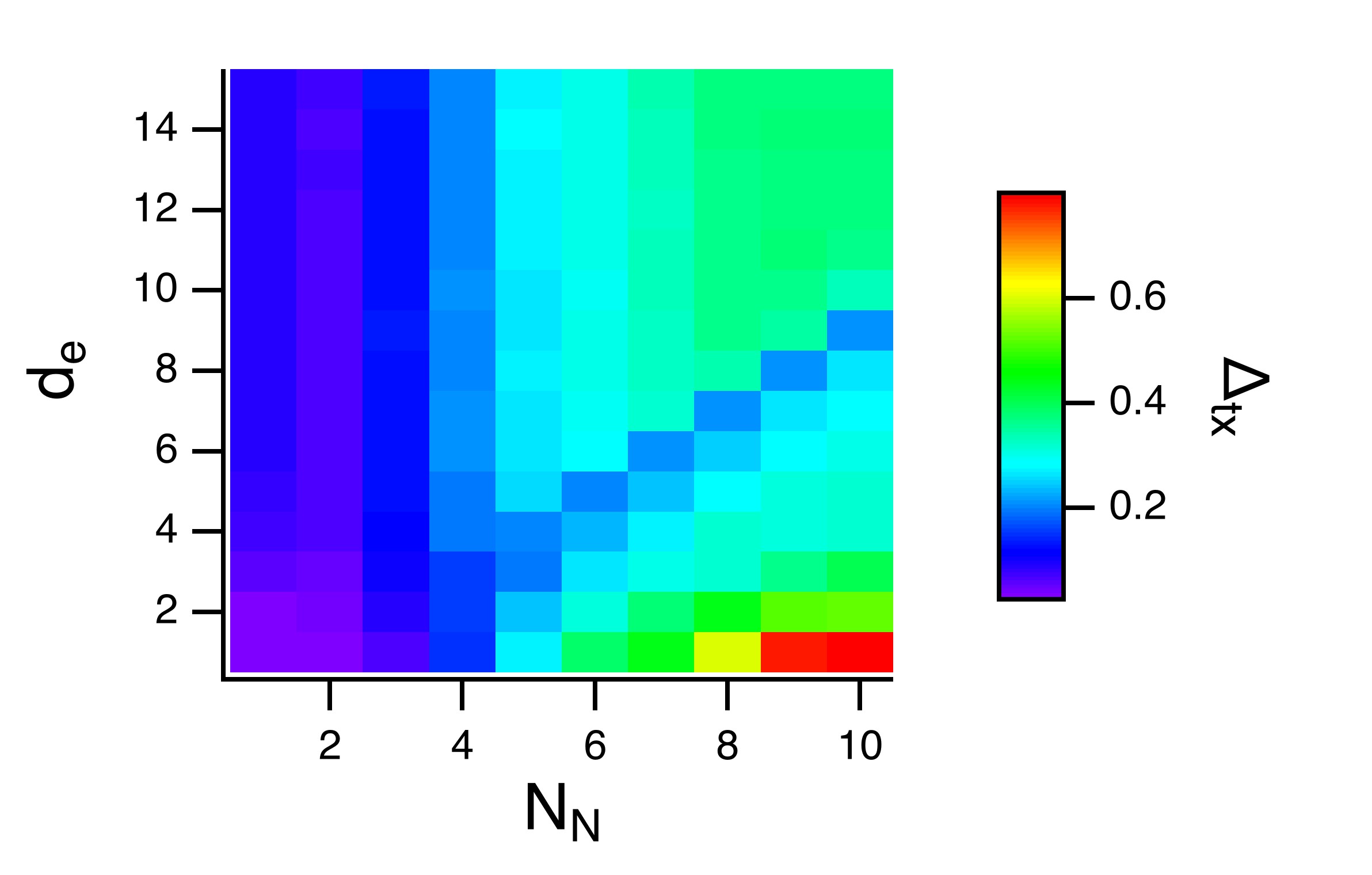} 
  \caption{ \label{vartanhvarnarma} Testing error for the multidimensional tanh nodes of eq. (\ref{ndtanh}) for the NARMA system of variable order $N_N$ as the node dimension $d_e$ varied.}
  \end{figure}

Figure \ref{vartanhvarnarma} shows the testing error when both the node dimension $d_e$ and the NARMA order $N_N$ are varied. For this plot, $g=0.35$ and $\varepsilon=0.35$ were chosen by varying both of these parameters for a NARMA system with $N_N=10$ and choosing the values that gave the lowest training error.

For NARMA systems with $N_N$ from 1 to 4, the lowest training error occurred for $d_e=1$. For higher order NARMA systems, the lowest training error was seen when $N_d=d_e-1$, demonstrating that the lowest training error came when the memory capacity for the reservoir matched the memory required by the problem being solved.

\section{Summary}
Memory is necessary for a reservoir computer, but there is an optimal amount of memory in a reservoir computer; too much or too little memory can lead to increased training errors. I have used Jaeger's standard definition of memory and I have introduced two alternate methods to measure memory; one that was suggested by \cite{inubushi2017} was based on using the variational equation for the reservoir computer to determine how the size of a perturbation changed with time, while the other, which was based on the ideas of capacity described in \cite{lymburn2020, jungling2021}, measured how the reservoir computers lost correlation over time.

It is necessary to tune the memory of a reservoir computer to get optimal results. This paper showed two methods; changing the sparsity of the adjacency matrix while maintaining the interaction strength between nodes, or constructing a multidimensional reservoir computer specifically to have more or less memory. Different measures of memory may be more or less useful for optimizing the reservoir computer for different problems. When building reservoir computers from analog hardware this sort of flexibility in choosing parameters may not always be available, so more work on designing reservoir computers to have a specified amount of memory is necessary. It would be useful in optimizing the reservoir computer to know beforehand how much memory was optimal for the particular problem. It is probable that systems whose autocorrelation drops quickly in time require less memory capacity than systems whose autocorrelation stays large for long times, but currently I know of no good way to quantify the optimal amount of memory short of simulating the full reservoir computer.

\section{Data Availability}
All necessary data is included in this paper.

\section{Acknowledgements}
I would like to thank Thomas Lymburn and Thomas J{\" u}ngling for useful discussions and feedback.

This work was supported by the Naval Research Laboratory's Basic Research Program.

\end{document}